\documentclass{article}
\usepackage{caption}
\usepackage{subcaption}
\usepackage{float}
\usepackage{hyperref}
\usepackage{amsmath}
\usepackage{multirow}
\usepackage{array}
\usepackage{booktabs}
\usepackage{adjustbox}
\usepackage{xcolor}

\usepackage{blindtext} % Package to generate dummy text throughout this template 

\usepackage[sc]{mathpazo} % Use the Palatino font
\usepackage[T1]{fontenc} % Use 8-bit encoding that has 256 glyphs
\usepackage{microtype} % Slightly tweak font spacing for aesthetics

\usepackage[english]{babel} % Language hyphenation and typographical rules

\usepackage[hmarginratio=1:1,top=32mm,columnsep=20pt]{geometry} % Document margins

\usepackage{enumitem} % Customized lists
\setlist[itemize]{noitemsep} % Make itemize lists more compact

\usepackage{abstract} % Allows abstract customization
\usepackage{titlesec} % Allows customization of titles
\renewcommand\thesection{\Roman{section}} % Roman numerals for the sections
\renewcommand\thesubsection{\roman{subsection}} % roman numerals for subsections
\titleformat{\section}[block]{\large\scshape\centering}{\thesection.}{1em}{} % Change the look of the section titles
\titleformat{\subsection}[block]{\large}{\thesubsection.}{1em}{} % Change the look of the section titles

\usepackage{fancyhdr} % Headers and footers
\usepackage{titling} % Customizing the title section

\usepackage{hyperref} % For hyperlinks in the PDF

\pretitle{\begin{center}\Huge\bfseries} % Article title formatting
	\posttitle{\end{center}} % Article title closing formatting
\title{Deep Tiling: Texture Tile Synthesis Using a Deep Learning Approach}
\author{%
	\textsc{Vasilis Toulatzis}\\[1ex] % Your name
	\normalsize University of Ioannina \\ % Your institution
	\normalsize \href{mailto:vtoulatz@cse.uoi.gr}{vtoulatz@cse.uoi.gr} % Your email address
	\and % Uncomment if 2 authors are required, duplicate these 4 lines if more
	\textsc{Ioannis Fudos}\\[1ex] % Your name
	\normalsize University of Ioannina \\ % Your institution
	\normalsize \href{mailto:fudos@cse.uoi.gr}{fudos@cse.uoi.gr}
}

\date{\today} % Leave empty to omit a date
\begin{document}

	% uncomment for using teaser
	% \teaser{
	%  \includegraphics[width=\linewidth]{eg_new}
	%  \centering
	%   \caption{New EG Logo}
	% \label{fig:teaser}
	%}
	
	\maketitle
	%-------------------------------------------------------------------------
	\begin{abstract}
		 Texturing is a fundamental process in computer graphics. Texture is leveraged to enhance the visualization outcome for a 3D scene. In many cases a texture image cannot cover a large 3D model surface because of its small resolution. Conventional techniques like repeating, mirror repeating or clamp to edge do not yield visually acceptable results. Deep learning based texture synthesis has proven to be very effective in such cases. All deep texture synthesis methods trying to create larger resolution textures are limited in terms of GPU memory resources. In this paper, we propose a novel approach to example-based texture synthesis by using a robust deep learning process for creating tiles of arbitrary resolutions that resemble the structural components of an input texture. In this manner, our method is firstly much less memory limited owing to the fact that a new texture tile of small size is synthesized and merged with the original texture and secondly can easily produce missing parts of a large texture.   
		
%		\printccsdesc   
	\end{abstract} 
	
	%\maketitle
	\section{Introduction}

Texture synthesis aims at generating a new texture such that its resolution and structure are instrumental in using it on wrapping a 3D model. Texture expansion plays a cardinal role in many applications where a large texture is necessary. Games along side with Geographic Information System (GIS) apps are such cases in which
large unbounded resolution textures are needed. 
%and constitute an essential contributory factor on main app's appearance.
 In addition, the same applies not only for diffuse textures but also for specular, normal, bump and height maps.

Structure similarity with the original input texture is one of the most investigated topics on texturing. Many texture synthesis methods aim at expanding a texture and usually on doubling its width and height. However, they simultaneously introduce a increase consumption of memory resources which severely restricts its scalability. To this end, many such methods end up on running in CPU without leveraging the power and speed of GPU. Consequently, memory efficiency is a key factor for texture synthesis and expansion.

Example-based texture synthesis techniques employ deep learning based optimization processes that seek larger resolution textures that resemble an input texture. Such methods by targeting on producing synthesized images of larger resolution textures do not have the capacity to create smaller or arbitrary resolution textures. Tiling is the only alternative for building arbitrary texture images. Therefore, tiling texture synthesis is not only capable of synthesizing larger textures but also a novel way of constructing step by step a brand-new texture or for completing missing parts of a larger texture. 
%giving us the ability of dropping and reconstructing a tile that does not fit our needs or purpose.

In this work, we propose a new texture synthesis approach that follows the aforementioned procedure. Thus, our method is capable of generating new tiles that match structurally and have the same morphology with the original input texture. We utilize a deep neural network to produce a new tile that can be used to expand the original texture. Subsequently our system builds a new texture of arbitrary shape and size by artificially synthesizing tiles in any direction.
	\section{Related Work}

\begin{figure*}[ht]
	\centering
	\includegraphics[width=0.95\textwidth]{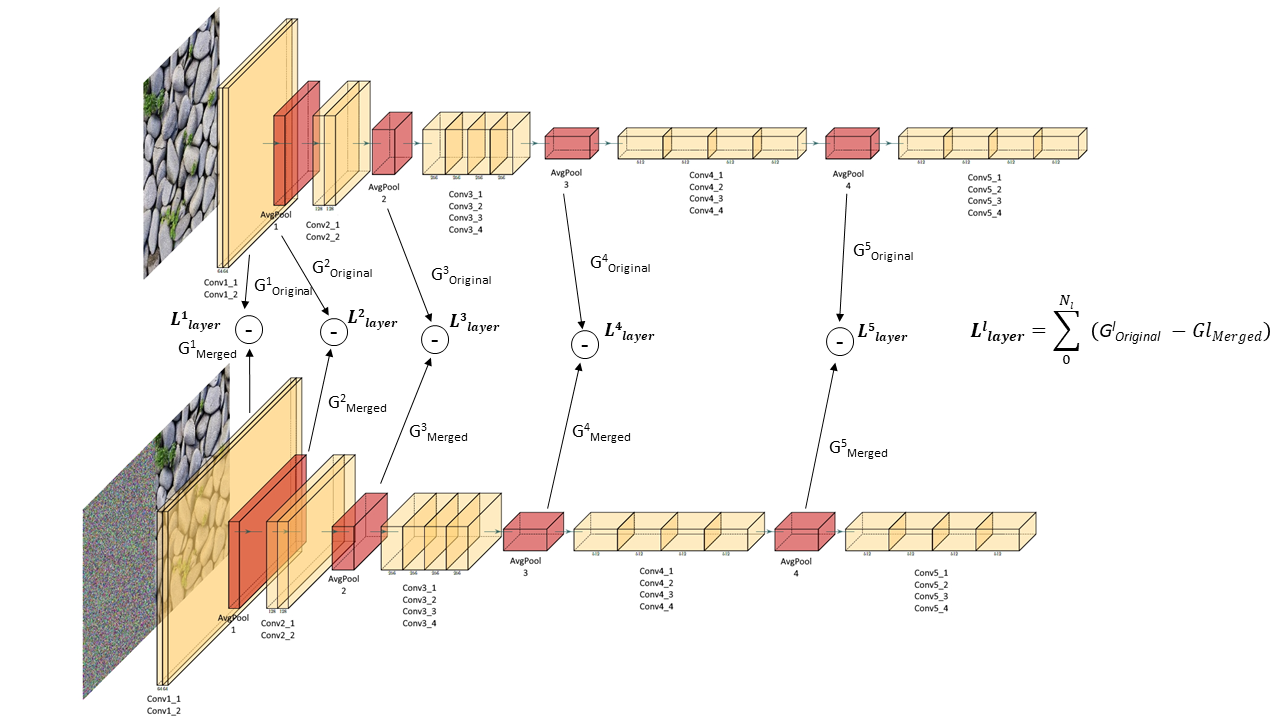}
	\caption{Deep texture tiling: $G{^l}_{layer}$ is the Gram Matrice of feature maps in layer $l$ which is merely dependent on number of filters ${N_l}$, network structure adopted is VGG19 \cite{vgg} but changing $Max Pooling$ layers to $Avg Pooling$ layers. This figure generated by PlotNeuralNet (https://github.com/HarisIqbal88/PlotNeuralNet) and then modified. }
	\label{fig:Method}
\end{figure*}

\subsection{Texture Synthesis}

Texture synthesis is a field of research that has drawn the attention of researchers for many years. Starting from simple ideas
of tiling patterns and stochastic models to state of the art techniques based on exemplars with all of them aiming to produce new synthesized and visually acceptable textures.

The most effective category are proved to be example-based methods that incorporate deep learning approaches \cite{Gatys:2015:TSU:2969239.2969269} (base of any other neural approach ahead of its time), optimization-based techniques \cite{Kwatra:2005:TOE:1073204.1073263}, pixel-based \cite{Efros:1999:TSN:850924.851569}, \cite{Wei:2000:FTS:344779.345009} and patch-based methods \cite{Kwatra:2003:GTI:1201775.882264}, \cite{Efros:2001:IQT:383259.383296}.

Expanding texture synthesis is the most challenging among the texture synthesis goals. Therefore, several techniques that aim at expanding  texture synthesis have been developed \cite{Kaspar:2015:STT:2816723.2816754}, \cite{Zhou:2018:NTS:3197517.3201285}. The most recent ones rely on deep learning by producing remarkable results on expanding and even for super-resolution texture synthesis \cite{sajjadi2017enhancenet}. By using Convolutional Neural Networks (CNNs) of many layers \cite{vc.20191262} or Generative Adversarial Networks (GANs) \cite{Fr_hst_ck_2019}, \cite{Zhou:2018:NTS:3197517.3201285} these methods correlate image features to produce a new synthesized high resolution texture map.

% The first one utilizes a Generative Adversarial Network (GAN) \cite{Goodfellow:2014:GAN:2969033.2969125}, trained to %expand a texture in a uniform manner. While, Self Tuning is a method extending the Texture Optimization %\cite{Kwatra:2003:GTI:1201775.882264}, \cite{Darabi:2012:IMC:2185520.2185578} and accomplishes visually acceptable %results. 

%-------------------------------------------------------------------------
\subsection{Texture Tiling}
The drawback of the aforementioned approaches is the large consumption of memory resources that makes them unsuitable for GPUs. To this end, tiling seems to be the only approach to synthesizing large textures. 
%until the GPU technologies will be able to lease their total power by upgrading to many GBs of RAM memory.
 
One of the simplest texture design techniques is repeating tile patterns such that the produced texture does not include seams. Moreover, methods generating stochastic tiles have been developed \cite{wangtiles03}, \cite{inproceedingswangtiles05} for the same texture synthesis purpose. %However, texture tiling is still an open field of research in terms of increasing texture resolution performance.

Thus, tiling forms a new challenge for texture synthesis and deep learning methods have already started being used to this end. 
%Their main advantage is the capability of synthesizing new textures that are not repeated. Instead, they are the very picture of the original input texture matching the corresponding original structure as follow up tiles in any direction.
One recent work that focuses on creating large tiles targeting on great resolution outcome texture is \cite{Fr_hst_ck_2019}. This work contributes
on homogenizing four texture tile outputs of GANs trained on lower resolution textures so as to produce a high resolution texture with no seam artifacts.

%-------------------------------------------------------------------------
%\subsection{Style Transfer}

%------------------------------------------------------------------------

	\section{Deep Texture Tiling}

We propose a novel algorithm of synthesizing tiles by using an alternative of the fundamental  work by \cite{Gatys:2015:TSU:2969239.2969269} on neural texture synthesis. In general, by leveraging the power of a CNN of multiple layers we extract and correlate feature maps across layers of two instances of a VGG19 \cite{vgg} network given two different resolution textures in each network. Our algorithm has the ability to synthesize texture tiles in a seamless manner by optimizing the distance of feature maps across the layers of our model by using as input textures the original and a new white noise tile merged with the original texture towards a specific orientation (up, down, right, left tiling).

Consequently, we embrace the main idea of deep texture synthesis but we abandon the specific size expansion and replace it with tiling. More specifically, we correlate feature spaces targeting to produce similar representations across network layers. On the first network we forward the original image while on the second network we utilize two user input tiling factors for width and height for a new white noise tile creation that is forwarded along with the original as a merged input texture.

For correlations among network layers extraction their feature space representations $F_{rc}^l$ of a general feature map $F^l\in R^{n_f \times vs_f}$ are needed, where $l$ is a layer having $n_f$ filters of size $vs_f$ reshaped into one dimension vectors. This is achieved by the use of Gram Matrices:

\begin{equation}
G_{rc}^l = \sum_{i} F_{ri}^l F_{ci}^l
\label{gram}
\end{equation}

The total layer loss is the sum of all layer losses that are computed as the mean squared displacement of the Gram Matrices of the two VGG19 instances. As a consequence, the total loss function is defined as follows:
\begin{equation}
L_{total}(I_{original}, I_{merged}) = \sum_{l=1}^{N^L} \frac{w^l}{4 {n^l_f}^2 {vs^l_f}^2}\sum ( G^{l}_{original} - G^{l}_{merged})^2
\label{lossfunction}
\end{equation}

where $I_{original}$ is the original texture and $I_{merged}$ is a white noise texture merged with the original one having been forwarded to our system as described above and $N^L$ the number of contributory layers. The whole process and the layers contributing to the total loss function are shown in Figure \ref{fig:Method}. We correlate feature representations along $layer1$, $pool1$, $pool2$, $pool3$, $pool4$ and the corresponding weight setting for every layer contributory factor is $\frac{1}{5}$.

n some texture input cases the output of our method produces some noise in the boundaries of the original and deep generated tile. Therefore, we developed an additional preprocessing phase we call Seam Removal in which we try to vanish the seam effect of deep texture tiling method. In the noise part of the second network instance ($Merged$ in Figure \ref{fig:Method}) instead of using a simple noise we utilize a mirrored version of the input original texture and then we apply noise that increases  exponentially as function of the distance of each column from the seam. Specifically, every pixel for the $Merged$ part of our model is computed as:

\begin{equation}
Noise(i, j) = w_1 Original(i, width-j-i) + w_2 RandomColor
\end{equation}

\begin{figure}
	\centering
	\begin{subfigure}[htb]{0.5\linewidth}
		\centering
		\includegraphics[width=0.86\textwidth]{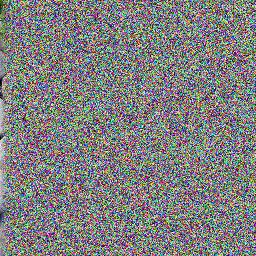}
		\caption{$a=0.25$}
		\label{subfig:newNoiseA}
	\end{subfigure}%    <-- % added here
	\hfill %% useful if width of each figure is less the .5\textwidth
	\begin{subfigure}[htb]{0.5\linewidth}
		\centering
		\includegraphics[width=0.86\textwidth]{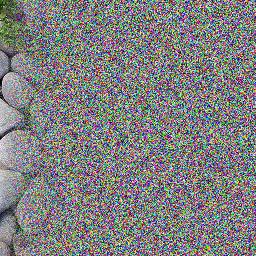}
		\caption{$a=0.05$}
		\label{subfig:newNoiseB}
	\end{subfigure}
	\caption{Seam Removal: exponential column mirroring}
\end{figure}

where $w_1 = e^{-\alpha j}$ with $\alpha\in(0, 1)$, $w_2 = 1 - w_1$, $i$ and $j$ rows and columns accordingly. In this manner, we make our method more robust and the produced noise outcome with $\alpha = 0.25$ and $0.05$ resembles like in Figures \ref{subfig:newNoiseA} and \ref{subfig:newNoiseB} respectively. An optimal $\alpha$ can been determined by $$\alpha=-\frac{50\,\,ln(0.5)}{c}$$ where $c\,x\,r$ is the resolution of the input texture. To obtain this we have determined experimentally that the optimal visual result is derived by setting as target an attenuation of 50\% (i.e. $w_1=0.5$) of the original mirrored image when we reach the 2\% of the total number of columns (i.e. $j=c/50$). By doing so we achieve a seamless join of the two images without the mirroring effect being noticeable.

%\begin{figure}[!tb]
%	\centering
%	\includegraphics[width=3cm]{Images/textureMirroredNoise.jpg}
%	\caption{Seam Removal: exponential column mirroring}
%	\label{newNoise}
%\end{figure}

	\section{Experiments \& Results}
\begin{figure*}[ht]
	\centering
	\includegraphics[width=\textwidth]{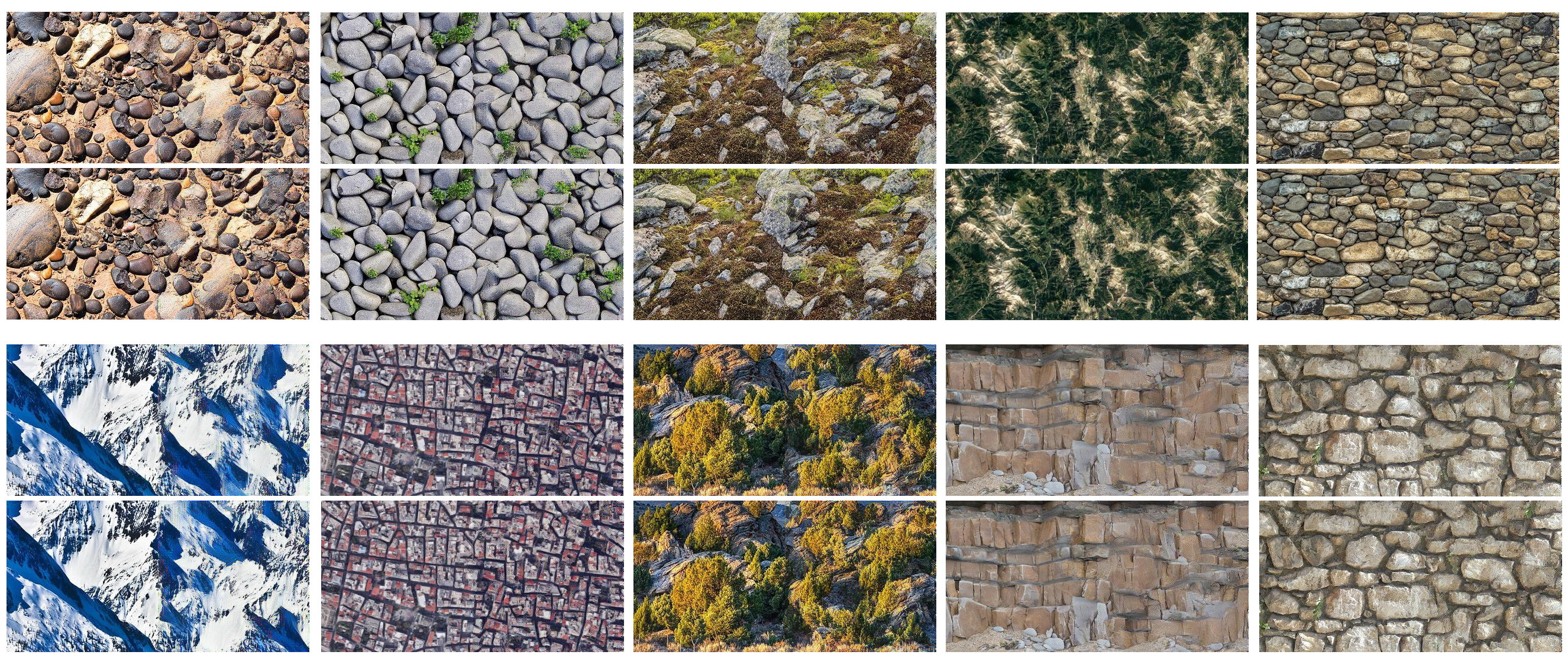}
	\caption{Results of deep right texture tiling: Odd lines have been generated by deep texture tiling for right tile construction without seam removal applied, even lines illustrate seam removal by exponential column mirroring with $\alpha = 0.15$.}
	\label{mainresults}
\end{figure*}

Our method has been developed on Python, using Tensorflow \cite{tensorflow} and it has been tested on an NVIDIA GeForce RTX 2080 Ti with 11GB GDDR6 RAM and 1350MHz base clock. Input texture resolution was $256 \times 256$ and we used a tiling factor of $1$ for both width and height of the generated input noise in both right and up tiling with which the original textures were merged (second VGG19 input). All outputs have been produced by $100000$ $iterations$ by utilizing the Adam optimizer \cite{adam} with $learning$ $rate = 0.0005$ on our system learning process and the running average time was $\approx2400$ $secs$.

\begin{figure}[!htb]
	\centering
	\includegraphics[width=0.9\linewidth]{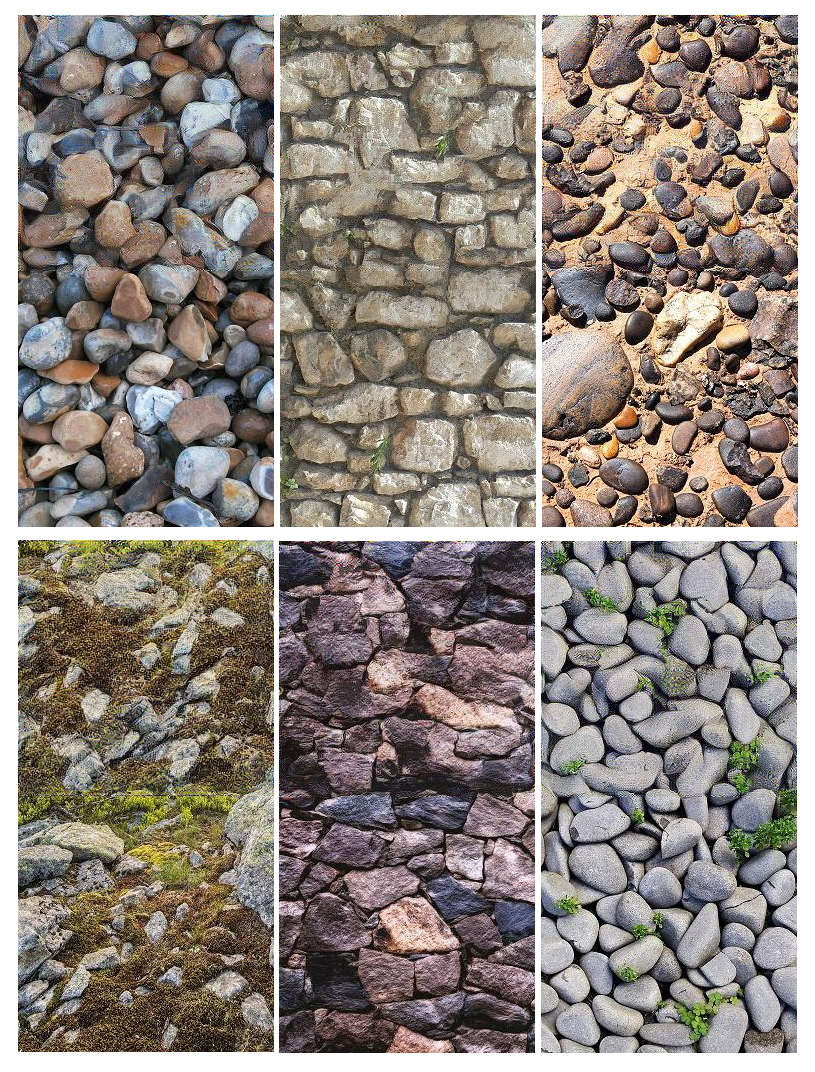}
	\caption{Results of deep up texture tiling: tiles synthesized on up direction with no seam removal.}             	\label{upTiling}
\end{figure}

In Figures \ref{mainresults} and \ref{upTiling} results of our algorithm are presented showing that synthesis of texture tiles which highly match an original texture is plausible by using our deep learning system with acceptable visual quality. Our method is able to be used in left and down tiling and in cases in which a part of texture is missing, as well. In latter cases, the merged texture in our work would be the union of all other tile surrounding the missing part.
	\section{Conclusions \& Future Work}

We presented an innovative tiling synthesis method that is capable of producing new texture tiles in any direction. Moreover, we introduce a new method for reducing seam effect in texture synthesis. Based on the results, our method has proven to be very effective on tile texture synthesis bearing an essential advantage for GPU texture synthesis execution due to the fact that tiles could be generated in small resolutions step by step, making even low RAM GPUs  capable of synthesizing high resolution textures. As future work, we are targeting at implementing and applying the same process in a GAN structure, expanding our method to style transfer by creating new tiles of mixed styles.
	%\input{Contents/Acknowledgement.tex}
	
	% bibtex
	\bibliographystyle{eg-alpha-doi} 
	\bibliography{egbibsample}       
	
\end{document}